%% file: main.tex
\documentclass[conference]{IEEEtran}

\usepackage[utf8]{inputenc}
\usepackage[T1]{fontenc}
\usepackage{microtype}
\usepackage{listings}
\usepackage{amsmath}
\usepackage[hidelinks]{hyperref}
\usepackage{cleveref}
\usepackage{multicol}
\usepackage{glossaries}
\usepackage[disable,commandnameprefix=always,todonotes={textsize=tiny,obeyFinal}]{changes}
\usepackage{tikz}
\usetikzlibrary{external}
\usepackage[binary-units=true]{siunitx}
\usepackage{dblfloatfix}
\usepackage{transparent}
\usepackage{bbm}
\usepackage{bm}
\usepackage{amssymb}
\usepackage{hyperref}
\usepackage{nicefrac}
\usepackage{booktabs}

\graphicspath{{figures/}}
\crefname{lstlisting}{listing}{listings}
\Crefname{lstlisting}{Listing}{Listings}
\crefname{figure}{Fig.}{Fig.}
\crefname{table}{Tab.}{Tab.}
\crefname{listing}{Listing}{Listing}

\usepackage[backend=biber,maxbibnames=6,minbibnames=6,mincitenames=1,firstinits=true,sorting=none,style=ieee,doi=true,eprint=false,isbn=false,url=false]{biblatex}
\usetikzlibrary{calc,backgrounds,fit,shapes.geometric}

\definecolor{color0}{HTML}{FA8D0F}
\definecolor{color1}{HTML}{0F69FA}

\DeclareMathOperator*{\argmin}{arg\,min}
\DeclareMathOperator*{\argmax}{arg\,max}

\newcommand{\ntap}{n_\textnormal{taps}}
\newcommand{\chunk}{\boldsymbol{y}^\text{c}[k]}

\input{acronyms}
\begin{document}
\title{Short-reach Optical Communications: 
\\A Real-world Task for Neuromorphic Hardware}
%
\author{%
	\IEEEauthorblockN{%
		Elias Arnold\IEEEauthorrefmark{1}\IEEEauthorrefmark{3}\textsuperscript{$\bigtriangleup$},
        Eike-Manuel Edelmann\IEEEauthorrefmark{2}\IEEEauthorrefmark{4}\textsuperscript{$\bigtriangleup$},
        Alexander von Bank\IEEEauthorrefmark{2}\textsuperscript{$\bigtriangleup$}\\
        Eric Müller\IEEEauthorrefmark{1},
        Laurent Schmalen\IEEEauthorrefmark{2}
        and
        Johannes Schemmel\IEEEauthorrefmark{1}}%
		\IEEEauthorblockA{\IEEEauthorrefmark{1}\textit{Kirchhoff Institute for Physics}, Heidelberg University, Germany}%
        \IEEEauthorblockA{\IEEEauthorrefmark{2}\textit{Communications Engineering Lab}, Karlsruhe Institute of Technology, Germany}%
		\IEEEauthorblockA{\IEEEauthorrefmark{3}\href{mailto:elias.arnold@kip.uni-heidelberg.de}{elias.arnold@kip.uni-heidelberg.de}}%
        \IEEEauthorblockA{\IEEEauthorrefmark{4}\href{mailto:edelmann@kit.edu}{edelmann@kit.edu}}%
        \IEEEauthorblockA{\textsuperscript{$\bigtriangleup$}contributed equally.}%
}

\maketitle

\begin{abstract}
\input{abstract}
\end{abstract}

\glsresetall


\input{introduction}

\input{imdd}

\input{previous}

\input{task}

\input{dataset}

\input{acknowledgments}

\AtNextBibliography{\footnotesize}
\printbibliography

\end{document}

%% file: acronyms.tex
\newacronym{snn}{SNN}{spiking neural network}
\newacronym{imdd}{IM/DD}{intensity-modulation, direct-detection}
\newacronym{fec}{FEC}{forward error correction}
\newacronym{rrc}{RRC}{root-raised-cosine}
\newacronym{pam4}{PAM-4}{pulse-amplitude-modulation 4-level}
\newacronym{ber}{BER}{bit error rate}
\newacronym{pd}{PD}{photodiode}
\newacronym{awgn}{AWGN}{additive white Gaussian noise}
\newacronym{isi}{ISI}{inter-symbol interference}
\newacronym{adc}{ADC}{analog-to-digital converter}
\newacronym{cd}{CD}{chromatic dispersion}
\newacronym{nn}{NN}{neuronal network}
\newacronym{dnn}{DNN}{deep neuronal network}
\newacronym{rnn}{RNN}{recurrent neuronal network}
\newacronym{motm}{MOTM}{max-over-time-membrane}
\newacronym{eotm}{EOTM}{end-over-time-membrane}
\newacronym{ttfs}{TTFS}{time-to-first-spike}
\newacronym{bss2}{BSS-2}{BrainScaleS-2}
\newacronym{pam}{PAM}{pulse-amplitude-modulation}
\newacronym{mlse}{MLSE}{maximum likelihood sequence estimator}
\newacronym{mmse}{MMSE}{minimum mean square error}
\newacronym{ann}{ANN}{artificial neural network}
\newacronym{led}{LED}{light emitting diode}
\newacronym{snr}{SNR}{signal-to-noise ratio}
\newacronym{lif}{LIF}{leaky integrate-and-fire}
\newacronym{lcdtask}{LCD-Task}{low chromatic dispersion-Task}
\newacronym{ssmftask}{SSMF-Task}{standard single-mode fiber-Task}
\newacronym{ml}{ML}{machine learning}
\newacronym{gpu}{GPU}{graphics processing unit}

%% file: abstract.tex
\Glspl{snn} emulated on dedicated neuromorphic accelerators promise to offer energy-efficient signal processing.
However, the neuromorphic advantage over traditional algorithms still remains to be demonstrated in real-world applications.
Here, we describe an \gls{imdd} task that is relevant to high-speed optical communication systems used in data centers.
Compared to other machine learning-inspired benchmarks, the task offers several advantages.
First, the dataset is inherently time-dependent, i.e., there is a time dimension that can be natively mapped to the dynamic evolution of \glspl{snn}.
Second, small-scale \glspl{snn} can achieve the target accuracy required by technical communication standards.
Third, due to the small scale and the defined target accuracy, the task facilitates the optimization for real-world aspects, such as energy efficiency, resource requirements, and system complexity.

%% file: introduction.tex
\section{Introduction}
The human brain is capable of complex signal processing with remarkably little energy consumption~\cite{padamsey2023paying}.
\Glspl{snn} mimic the behavior of the human brain, where, unlike in \glspl{ann}, neurons and synapses exhibit time-dependent dynamics and communicate information in an event-based way via discrete temporal events\chadded{,} called spikes.
To realize fast low-power information processing, neuromorphic hardware accelerators emulate brain-inspired \glspl{snn} to adopt the brain's underlying principles of event-based computing.

In recent years, the parameter space of state-of-the-art \glspl{dnn} has grown enormously~\cite{maslej2024artificial}.
To demonstrate the potential benefits of event-based modeling approaches and neuromorphic hardware in particular, it is rather problematic to use them as benchmarks, if only for reasons of hardware size requirements and cost~\cite{thakur2018mimicthebrain_nourl}.
Consequently, it is challenging to compare \gls{snn}-based solutions to traditional approaches on commonly used \gls{ml} datasets.
The evaluation of \glspl{snn} on small hardware substrates is often driven by toy-like datasets, that either have no real-world application, do not exhibit temporal structure, or lack a reasonable amount of data.
Datasets that allow evaluation of these systems on a real application with the potential to demonstrate a neuromorphic advantage are scarce.

Accompanying hardware development, the development of training and learning algorithms for \glspl{snn}, and the efficient neural coding of real-world data signals and other information, such as the mapping of temporal real-valued data series into spike-based representations, are still current research topics~\cite{guo2021neural}.
Hence, there is a need for datasets with an emphasis on neural resource efficiency and sparse input encoding, exhibiting natural temporal structure, and allowing the evaluation of the efficiency of algorithms for temporal information processing~\cite{yik2023neurobench}.

Commonly used datasets for testing and evaluating small-scale neuromorphic substrates, as well as optimizing training algorithms for \glspl{snn}, comprise the MNIST dataset \cite{lecunmnist}, or its neuromorphic version N-MNIST \cite{orchard2015converting}, the Yin-Yang dataset \cite{kriener2021yin}, and the Spiking Heidelberg Digits dataset \cite{cramer2022heidelberg,cramer2022surrogate}.
Other neuromorphic datasets, like DVS128 Gesture \cite{lowpower2017Amir} or CIFAR10-DVS \cite{cifar10dvs2017li}, exist, but tend to require larger \gls{nn} topologies.

With this work, we contribute an optical communication dataset generator and two pre-defined parameterizations to the broader \gls{ml} community, compelling for advancing neuromorphic hardware and algorithm development with a real-world application in mind.

In communication engineering, data is transmitted via a time-varying signal.
The demand for higher data rates necessitates increasingly powerful and efficient transmitters and receivers.
As a result, these processing stages get more power-hungry, increasing the overall energy consumption of communication systems.
Hence, there is a need for novel transmitters and receivers that are both, high-performing and energy-efficient.
Communication signals inherently have a time dimension:
The transmitter sends signals to the receiver over a physical transmission medium, the \emph{channel}, which disturbs the transmission.
The effects of channels can be well approximated by mathematical models, allowing for accurate channel simulations. 
In high-speed non-coherent optical communications, \gls{imdd} models with \gls{cd} are crucial for modeling optical interconnects in passive optical networks, data centers and system-on-chip interconnects~\cite[p.122]{handbook_optical_networks}.
In \gls{imdd} systems, increasing data rates intensify the \gls{cd}, increasing the need for additional signal processing as a countermeasure. 
To this end, equalizers are applied at the receiver side, enabling more reliable transmission. 
Demappers finally map the equalized signal to an estimate of the transmit data (bits).
In~\cite{BankSPPCom,vonBank2024energy,arnold2022spiking,arnold2023spiking,arnold2022spikinghardware}, \gls{snn}-based equalizers and demappers for compensating impairments in a simulated \gls{imdd} link were demonstrated.
\Gls{snn}-based approaches have shown superior performance compared to linear and \gls{ann}-based approaches.
Notably, in~\cite{arnold2022spikinghardware} a \gls{snn} for joint equalization and demapping was successfully implemented on the analog \gls{bss2} platform, meeting the required \gls{ber} of the communication system.
In future hardware, received samples could be encoded directly into spike events, allowing receivers to operate fully in analog without requiring energy-hungry digitization.

The practical implementation of an \gls{imdd} link emphasizes minimal resource and energy requirements at the transmitter and receiver side, making it particularly suitable for small-scale systems with low-power information processing algorithms.
For reliable data transmission, \glspl{ber} must remain below a well-defined threshold while the receiver maintains a demanded throughput, thereby defining clear objectives for the processing device.
The intrinsic temporal dimension of communication signals offer opportunities to explore algorithms that leverage temporal information, such as synaptic delays, adaptation mechanisms, or recurrence.
Additionally, potential solutions may benefit from efficient neural coding of the received input signal.
We provide a PyTorch-based dataset generator simulating the \gls{imdd} link publicly\footnote{\url{https://github.com/imdd-task/imdd-task}} and detail its usage.
Simulating the link allows for the generation of unlimited data, enabling extensive testing and optimization.
We provide the reader with commonly used methods for training \glspl{nn}-based receivers and outline its intricacies.
Finally, we define two \gls{imdd} link parameterizations as different tasks, a \gls{lcdtask} and a realistic \gls{ssmftask}, for which we present baseline results using \gls{snn}-based receivers.

%% file: imdd.tex
\section{Optical Communications} \label{sec:imddmodel}

\subsection{Communication Problem}
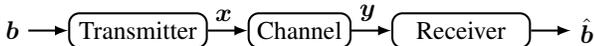
\begin{figure}[htb]
    \centering
    \resizebox{.45\textwidth}{!}{\input{comm_sys}}
    \caption{Sketch of a communication system.}
    \label{fig:comm_sys}
\end{figure}

\Cref{fig:comm_sys} shows the general setup of a communication system.
Its goal is to transmit a binary message $\bm{b}\in\{0,1\}^{MN}$ of length $MN$ from a transmitter to a receiver over a physical channel without error.
To transmit the binary message, a group of ${M=\log_2|\mathcal{X}|}$ bits is mapped to a transmit symbol ${x\in\mathcal{X}}$, where ${\mathcal{X} \subset \mathbb{R}}$ denotes the set of transmit symbols\footnote{In general, the transmit symbols are complex, i.e., ${\mathcal{X} \subset \mathbb{C}}.$} and the mapping is a bijective function between bit patterns and transmit symbols.
Hence, transmission of $M \cdot N$ bits results in a sequence of $N$ transmits symbols ${\bm{x} \in \mathcal{X}^N}$.
The transmit symbols $\bm{x}$ are disturbed by the channel, resulting in the channel output ${\bm{y} \in \mathbb{C}^N}$. 
Based on the observation of~$\bm{y}$, the receiver outputs an estimate of the transmit bit sequence~$\hat{\bm{b}}$.

A common metric to classify the performance of a communication system is the \gls{ber}.
It is defined as the average bit-level mismatch between $\bm{b}$ and $\hat{\bm{b}}$,
\begin{align}
    \mathrm{BER} = \frac{1}{MN} \sum_{n=0}^{MN-1} \mathbbm{1}\left( b_n \neq \hat{b}_n \right)\, ,
\end{align}
with 
\begin{align}
    \mathbbm{1}\left( b_n \neq \hat{b}_n \right) = \begin{cases}
        0 \quad &\text{if} \quad b_n = \hat{b}_n\, , \\
        1 \quad &\text{if} \quad b_n \neq \hat{b}_n \\
    \end{cases} \, .
\end{align}
To enable an error-free transmission, the receiver typically consists of different stages of signal processing and error correction.
Hereby, each stage further reduces the \gls{ber}. 

\subsection{Intensity Modulation with Direct Detection (IM/DD)}

\begin{figure}[hb]
    \centering
    \resizebox{.49\textwidth}{!}{\input{imdd_channel}}
    \caption{\label{fig:imdd_sketch}%
    Simplified setup of an \gls{imdd} link.
    }
\end{figure}
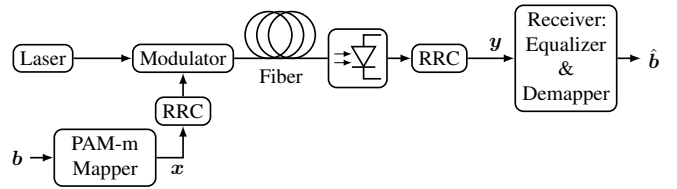

In non-coherent optical communications, intensity modulation with direct detection is often used.
The simplified setup of an \gls{imdd} link can be seen in~\Cref{fig:imdd_sketch}.
Again, $M$ bits are grouped together and mapped to a real-valued transmit symbol using a \gls{pam}, where each of the ${m=2^M}$ symbols of the \gls{pam} represents an amplitude level.
Using a \gls{rrc} filter with roll-off factor $\beta$ as pulse shaping filter, the transmit sequence $\bm{x}$ is converted into a continuous-time signal, which is modulated onto the laser intensity.
Then, the modulated light is coupled to the optical fiber.
Inside the fiber, the frequency-dependent propagation speed results in \gls{cd}, and the symbols experience dispersion in the time domain. 
Hence, consecutive symbols overlap and interfere, yielding the so-called \gls{isi}.
The impact of \gls{cd} on the $k$-th received symbol $y[k]$ can be narrowed down to the $n_\mathrm{ISI}$ neighboring symbols.
The strength of \gls{isi} depends on the length of the fiber, its dispersion coefficient, laser wavelength, and the data rate.
At the output of the optical fiber, the signal intensity is measured by a \gls{pd}, corresponding to a squaring of the electric field and introducing a non-linear distortion.
Additionally, thermal noise in the trans-impedance amplifier following the \gls{pd} is modeled as \gls{awgn} with variance $\sigma_\mathrm{n}^2$.
In summary, we can write
\begin{align}
    y[k] = &f\left( x\left[k - \Big\lfloor \frac{n_\mathrm{ISI}}{2} \Big\rfloor \right],\ldots,x[k],\ldots, x\left[k + \Big\lfloor \frac{n_\mathrm{ISI}}{2} \Big\rfloor \right] \right) \\ \nonumber
    &+ n[k]\, , \qquad n[k] \sim \mathcal{N}(0,\sigma_\mathrm{n}^2)  \, ,
\end{align}
where $f(\cdot): \mathcal{X}^{n_{\mathrm{ISI}}}\rightarrow \mathbb{R}$ is an arbitrary function modeling both \gls{isi}~\cite[p.~628]{proakis2008} and the non-linear distortion; $n[k]$ models the \gls{awgn}.

After the \gls{pd}, another \gls{rrc} filter is applied and the sequence of received values $\bm{y}$ is sampled over time.
The sequence $\bm{y}$ is passed to the receiver, which implements both channel equalization and demapping.
The receiver tries to mitigate the impact of the \gls{imdd} link and recover the transmit bits $\bm{\hat{b}}$.

To realistically model the continuous-time \gls{imdd} link using discrete-time simulations, oversampling ---as used in~\cite{arnold2023spiking}--- is required.
For a more detailed description of the implementation of the \gls{imdd} link, the interested reader is referred to~\cite{arnold2023spiking}.

\subsection{PAM-4 Mapper}
In this work, a \gls{pam4} is used.
Thus, $m=4$ intensity levels exist, and pairs of $M=\log_2 m = 2$ bits are mapped into one transmit symbol.
We denote the bits mapped to $x[k]$ as $b_1[k]$ and $b_2[k]$.
Gray mapping is used.
\Cref{tab:pam4_mapping} shows the mapping from bits to the symbol index $q$.
Both the bits and the transmit symbols are independent and identically distributed.
Before transmission, the transmit symbols are normalized to an average power of~$P_\mathrm{avg}=1$.

\begin{table}[h!]
    \vspace{-\baselineskip} 
    \centering%
    \caption{\label{tab:pam4_mapping}%
    \Gls{pam4} Mapping%
    }%
    \begin{tabular}{l cccc}
         \toprule
         Symbol index $q$ & 1 & 2 & 3 & 4 \\
         \midrule
         $b_1[k]\,b_2[k]$ & \texttt{00} & \texttt{01} & \texttt{11} & \texttt{10}\\
         \bottomrule
    \end{tabular}
\end{table}

\subsection{Practical Requirements}

To ensure reliable transmission in practical systems, at the output of the receiver, a target \gls{ber} is defined.
For a given link and noise power $\sigma_\mathrm{n}^2$, more powerful receivers can achieve a lower \gls{ber}, resulting in a greater margin to the target \gls{ber}.
This margin can be used to either realize higher data rates, transmit over longer fibers, or use less transmit power.

%% file: comm_sys.tex
\begin{tikzpicture}[>=latex, thick, rounded corners]
\def \xdist{.55cm}
    \node[] (x) {$\bm{b}$};
    \node[draw,rectangle,right=\xdist of x,minimum width = 1.9cm] (tx) {Transmitter};
    \node [right= \xdist of tx, draw, rectangle] (ch) {Channel};
    \node[right = \xdist of ch, draw, rectangle,minimum width = 1.9cm] (rx) {Receiver};
    \node[right=\xdist of rx] (y) {$\hat{\bm{b}}$};

    \draw[->] (x) -- (tx);
    \draw[->] (tx) -- (ch) node[above, pos=.4] {$\bm{x}$};
    \draw[->] (ch) -- (rx) node[above,pos=0.4] {$\bm{y}$};
    \draw[->] (rx) -- (y);
\end{tikzpicture}
    

%% file: imdd_channel.tex
\begin{tikzpicture}[>=latex, thick]
    \def\xdist{.4cm}

    \node[](b){$\bm{b}$};
    \node[draw, rectangle, right=\xdist of b, rounded corners,text width=1.5cm, align=center] (map) {PAM-m \\ Mapper};
    \node[above right=1.2cm and -.3cm of b, draw, rectangle, rounded corners] (laser) {Laser};
    \node[right=2.5*\xdist of laser,draw, rectangle, rounded corners] (mod) {Modulator};
    \node[below=.4cm of mod, draw, rectangle, rounded corners] (rrc) {RRC};
    \node[right=4*\xdist of mod, draw,rectangle, rounded corners,minimum height =1cm, minimum width=1cm] (diode){};
    \node[right=\xdist of diode, draw, rectangle, rounded corners] (mf) {RRC};
    \node[right=2*\xdist of mf, draw, rectangle,text width=1.5cm,align=center, rounded corners] (eq) {Receiver: \\
    Equalizer \& \\ Demapper};
    \node[right=\xdist of eq] (b_out) {$\hat{\bm{b}}$};

    \draw[->] (b) -- (map);
    \draw[->] (map) -| (rrc) node[pos=.4, below] {$\bm{x}$};
    \draw[->] (rrc) -- (mod);
    \draw[->] (laser) -- (mod);
    \draw[->] (diode) -- (mf);
    \draw[->] (mf) -- (eq) node[pos=.6,above] {$\bm{y}$};
    \draw[->] (eq) -- (b_out);
    
    \draw[thick] (mod) -- (diode);

    \def\off{10*\xdist}
    \foreach \x in {\off+0.3cm,\off+0.5cm,\off+0.7cm} {
    \draw[thick] (\x,2.1cm) circle(0.4) ;
    }
    \node[below right=-.2cm and .3cm of mod] () {Fiber};

    \draw[-] (diode)+(-.1,.2) -- ++(.3,.2) -- ++(-.2,-.34) -- cycle;
    \draw[-] (diode)+(-.1,-.15) -- ++(.3,-.15);
    \draw[-] (diode)+(.1,.2) -- ++(.1,.4) -- ++(.3,0);
    \draw[-] (diode)+(.1,-.15) -- ++(.1,-.35) -- ++(.3,0);
    \draw[->,thin] (diode)+(-.4,.075) -- ++(-.1,.075);
    \draw[->,thin] (diode)+(-.4,-.075) -- ++(-.1,.-.075);
\end{tikzpicture}

%% file: previous.tex
\section{Previous Work}
To compensate \gls{isi}, linear equalizers such as the \gls{mmse} equalizer~\cite[pp.~645-646]{proakis2008} can be deployed, offering a low complexity~\cite[p.~640]{proakis2008}.
However, for the \gls{imdd} link, the low complexity comes at the cost of an inability to address non-linear impairments~\cite{arnold2022spikinghardware}.
To implement non-linear equalization and demapping, \gls{ann}-based approaches have been explored in, e.g.,~\cite{bluemn23_annequalizer}.
\Glspl{snn} have also been studied for joint equalization and demapping of the \gls{imdd} link~\cite{arnold2023spiking,boecherer2023spiking,li2024spiking,vonBank2024energy}, demonstrating performance comparable to or even exceeding \gls{ann}-based receivers.
This makes them promising candidates for low-power signal processing when implemented on dedicated neuromorphic hardware.

To estimate a pair of $M$ bits, for \gls{pam4} $b_1[k]$ and $b_2[k]$, the \glspl{snn} in these studies process chunks of samples $\chunk$.
The chunk $\chunk$ is an excerpt of $\bm{y}$ of length $n_\mathrm{taps}$, centered around $y[k]$, where $n_\mathrm{taps}$ defines the number of received symbols required for equalization and demapping:
\begin{align}
    \chunk = \left[y\left[k - \Big\lfloor \frac{n_\mathrm{taps}}{2} \Big\rfloor \right],\ldots,y[k],\ldots, y\left[k + \Big\lfloor \frac{n_\mathrm{taps}}{2} \Big\rfloor \right] \right].
\end{align}
As the full required information is assumed to be contained within $\chunk$, chunks are processed independently and in parallel.
Hence, the \glspl{snn} do not maintain state across chunks.
It is noteworthy that by using chunks, the \glspl{snn} cannot exploit the intrinsic time dimension of the data.

Challenges arise in both translating $\chunk$ into a spiking representation that can be processed by the \gls{snn} and decoding the output of the \gls{snn} into bits $\bm{\hat{b}}$.
While \cite{arnold2023spiking} and \cite{li2024spiking} used a spike-timing approach, where each sample in $\chunk$ is scaled to spike times across a population of neurons representing its value,~\cite{BankSPPCom} converts each sample into a bit pattern using a quantizer, where each quantization bit is linked to an input neuron, determining its firing.
In~\cite{vonBank2024energy}, a real-valued encoding is proposed, aiming at a joint optimization of receiver performance and generated spikes. 
Various decoding strategies, such as \gls{motm}, \gls{eotm}, \gls{ttfs}, and rate-based methods, have been compared in \cite{li2024spiking}.
While \gls{motm} and \gls{eotm} rely on non-spiking leaky-integrator neurons, \gls{ttfs} and rate decoding utilize spiking output neurons.
In \cite{arnold2023spiking}, a spiking receiver was successfully deployed on the analog neuromorphic \gls{bss2} system \cite{pehle2022brainscales2,mueller2022scalable}, highlighting its feasibility for small-scale low-power analog signal processing.

\subsection{Learning Methods}
In the following, we provide the reader with a brief overview of equalization and demapping approaches.
Note that this is not extensive and does not cover learning methods that work on the intrinsic time dimension of the data.

\subsubsection{Equalization + Hard Decision}

Inspired by the linear \gls{mmse} equalizer~\cite[~pp.~645--646]{proakis2008}, an parameterized model outputs an estimate $\tilde{x}[k] \in \mathbb{R}$ based on a chunk $\chunk$. 
By determining the minimal Euclidean distance, a hard decision $\hat{x}[k] = \argmin_{x\in\mathcal{X}} |\tilde{x}[k] - x|$ converts the estimate into an estimation of the transmit symbol $\hat{x}[k]$.
The parameters of the model can be determined by, e.g., the \gls{mmse} loss of $\tilde{x}[k]$ and $x[k]$.
The bits are obtained by inverting the respective mapping, e.g., using \cref{tab:pam4_mapping}.

\subsubsection{Symbol-level Equalization and Demapping}
\label{sec:symbol_level}

Equalization and demapping on a symbol level is approached in~\cite{arnold2023spiking,boecherer2023spiking,li2024spiking,vonBank2024energy} and can be described as a classification task of the symbols: Given $\chunk$, a parameterized model returns the probabilities ${P_{\hat{X}}(\hat{x}[k]=x),\, \forall x \in \mathcal{X}}$.
For, e.g., an \gls{snn}, $|\mathcal{X}|$ output neurons can be used, each corresponding to a possible transmit symbol~$x$.
To obtain the probabilities, the outputs of the \gls{snn} are fed to a softmax function.
The model can be optimized using, e.g., the cross-entropy loss.
During application, the transmit symbol is  hard decided by $\hat{x}[k] = \argmax_{x \in \mathcal{X}} P_{\hat{X}}(\hat{x}[k]=x)$.
Again, the bits are obtained by inverting the respective mapping, e.g., using \cref{tab:pam4_mapping}.

\subsubsection{Bit-level Equalization and Demapping}

The previous approaches return hard-decided bits. 
However, by returning the probability of a bit being either zero or one, so-called soft-bits, the overall communication system can benefit from the use of soft-decision \gls{fec} codes~\cite[pp.~436--438]{proakis2008}.
Chunks are processed to estimate the $i$-th bit ${b_i,\, i\in\{1,\ldots,M\}}$, a parameterized model outputs the log-likelihood ratio 
\begin{align} 
    \ell_i = \log \left(\frac{P_{B_i}(B_i=1)}{P_{B_i}(B_i=0)}\right) \, .
\end{align}
Hence, $M$ outputs are needed.
The model can be optimized using, e.g., the binary cross-entropy loss.
The soft bits can be passed to the next stage of the communication system.
To hard decide the $i$-th bit, $\hat{b}_i = \frac12\left(1 + \text{sign}(\ell_i) \right)$ can be applied.

%% file: task.tex
\section{Tasks}\label{sec:task}
\subsection{Parameterization}
To optimize and evaluate novel receivers, we provide an implementation of the \gls{imdd} link:
The parameters are the number \texttt{N} of transmit symbols, the length \texttt{n\_taps} of a chunk $\chunk$, the set of transmit symbols \texttt{alphabet}, the \texttt{oversampling\_factor}, the \texttt{baudrate}, the \texttt{wavelength} of the laser, the \texttt{dispersion\_parameter} determining the strength of the fiber's \gls{cd}, the \texttt{fiber\_length}, the \texttt{noise\_power\_dB}, the \texttt{roll\_off} factor $\beta$ of the \gls{rrc} filters, and a \texttt{bias}, which is added in the simulation after pulse shaping to ensure a positive-only output of the \gls{rrc}.

\subsection{Parameter Sets}
In the following, we define two different parametrizations of the \gls{imdd} link as separate tasks.
The first \gls{lcdtask} emphasizes non-linear impairment with little \gls{cd} and thus weak \gls{isi}.
Consequently, receivers require a smaller number of $\ntap$ consecutive input samples or shorter memory, making it an excellent dataset for exploring resource efficiency and novel algorithms with small-scale networks, possibly implemented on small-scale prototype hardware.
The parameters of the \gls{lcdtask} are shown in~\Cref{tab:params_task1}.

\begin{table}[tbh]
    \centering
    \caption{Parameter Set: \acrfull{lcdtask}}
    \label{tab:params_task1}
    \begin{tabular}{l c}
        \toprule
        \textbf{Parameter} & \textbf{Value} \\
        \midrule
        \texttt{N} & \num{10000} \\
        \texttt{n\_taps} & \num{7} \\
        \texttt{alphabet} & $\left[-3, -1, 1, 3\right]$  \\
        \texttt{oversampling\_factor} & \num{3}  \\
        \texttt{baudrate} & $112\,\mathrm{GBd}$  \\
        \texttt{wavelength} & \SI{1270}{\nano\meter}  \\
        \texttt{dispersion\_parameter} & \SI{-5}{\pico\second \per \nano\meter \per \kilo\meter}  \\
        \texttt{fiber\_length} & \SI{4}{\kilo\meter}  \\
        \texttt{noise\_power\_db} & \SI{-20}{\dB}  \\
        \texttt{roll\_off} & \num{0.2}  \\
        \texttt{bias} & \num{2.25}\\
        \bottomrule
    \end{tabular}
\end{table}

The second \gls{ssmftask} models the standard single-mode fiber with a used wavelength of $\SI{1550}{\nano \meter}$, which is widely used in practice due to its low attenuation and, thus, wide range.
However, the \gls{imdd} link suffers from severe \gls{isi}, increasing the number of required input samples $\ntap$ and hence the need for more powerful equalizers.
\Cref{tab:params_task2} shows the parameters of the \gls{ssmftask}.

\begin{table}[tbh]
    \centering
    \caption{Parameter Set: \acrfull{ssmftask}}
    \label{tab:params_task2}
    \begin{tabular}{l c}
        \toprule
        \textbf{Parameter} & \textbf{Value} \\
        \midrule
        \texttt{N} & \num{10000} \\
        \texttt{n\_taps} & \num{21} \\
        \texttt{alphabet} & $\left[0, 1, \sqrt{2}, \sqrt{3}\right]$  \\
        \texttt{oversampling\_factor} & \num{3}  \\
        \texttt{baudrate} & $50\,\mathrm{GBd}$  \\
        \texttt{wavelength} & \SI{1550}{\nano\meter}  \\
        \texttt{dispersion\_parameter} & \SI{-17}{\pico\second \per \nano\meter \per \kilo\meter }  \\
        \texttt{fiber\_length} & \SI{5}{\kilo\meter}  \\
        \texttt{noise\_power\_db} & \SI{-20}{\dB}  \\
        \texttt{roll\_off} & \num{0.2}  \\
        \texttt{bias} & \num{0.25}\\
        \bottomrule
    \end{tabular}
\end{table}

\begin{figure}[tbh]
    \centering
    \input{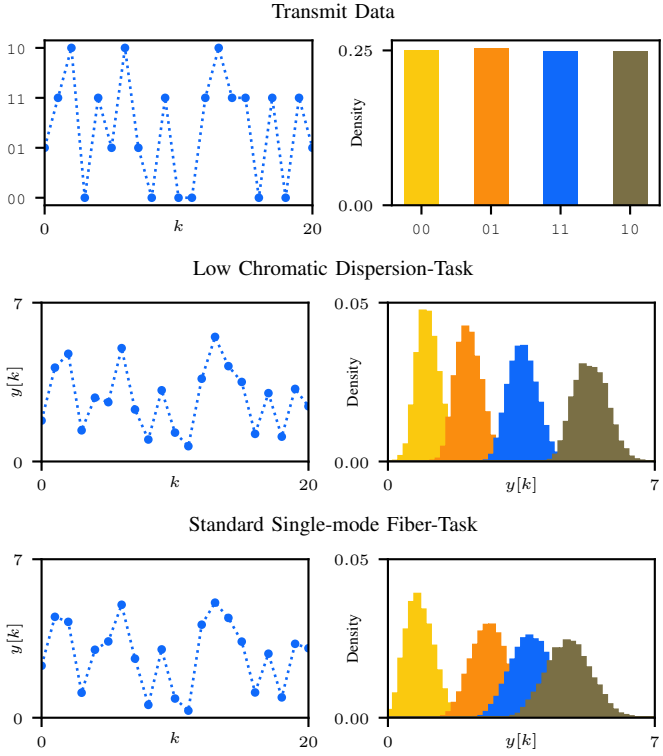}
    \vspace{-20px}
    \caption{Exemplary transmit and receive data for the \gls{lcdtask} and \gls{ssmftask}.
    The left column visualizes the time-discrete transmit, respective received signal, while the right column visualizes the corresponding histograms.
    Both links were simulated at $\sigma_\mathrm{n}^2=-\SI{20}{\decibel}$.}
    \label{fig:imdd_link_data}
\end{figure}

\Cref{fig:imdd_link_data} visualizes the time-discrete signals of the transmit and receive data for both tasks (left column) as well as the corresponding histograms (right column).
In the upper row, the transmit signal with the four amplitude levels and its histogram is visualized.
The middle and bottom rows show the received signals and the corresponding histograms for the \gls{lcdtask} and \gls{ssmftask}, respectively.
Due to stronger \gls{isi} of the \gls{ssmftask}, the symbol distributions of the received samples in dependence of the transmit symbols exhibit more overlap.

During training, the user is free to alter the parameters, e.g., the noise power $\sigma_\mathrm{n}^2$. 
However, during testing, only \texttt{n\_taps} can be chosen freely.

\subsection{Evaluation}

To demonstrate the demapping performance, the achieved \gls{ber} at a noise power of $\sigma_\mathrm{n}^2=\SI{-20}{\decibel}$ should be reported.
Furthermore, the \gls{ber} for various $\sigma_\mathrm{n}^2$ should be plotted, showing the ability of the receiver to generalize over a wide range of $\sigma_\mathrm{n}^2$. 
Incrementing the noise power $\sigma_\mathrm{n}^2$ by $\SI{1}{\decibel}$ provides a sufficiently fine resolution.
To ensure statistical significance of the reported \gls{ber}, for each $\sigma_\mathrm{n}^2$ and thus \gls{ber} data point, new data should be transmitted until a minimum of \num{2000} bit error events are encountered.

To measure the complexity of the model and anticipate its energy consumption, the number of trainable parameters and, in the case of \glspl{snn}, the average number of spikes sent to the \gls{snn} and the spikes emitted by the neurons within the \gls{snn} per processing of one symbol are to be reported.

As this task focuses on algorithmic and hardware development, we refrain from reporting processing speed, even though throughput is an important requirement for actual hardware realizations in communication systems and, if available, should be added.

\subsection{Results}

\begin{figure*}[tbh]
    \centering
    \tikzset{
        panel/.style={
            inner sep=0pt,
            outer sep=0pt,
            execute at begin node={\tikzset{anchor=center, inner sep=.33333em}}},
        label/.style={
            anchor=north west,
            inner sep=0,
            outer sep=0}}

    \begin{tikzpicture}
        \node[panel, anchor=north west] (a) at (0,  0) {
            \input{snr_task1.pgf}};
        \node[label] at (0, 0) {\textbf{A}};
        \node[panel, anchor=north west] (b) at (9,  0) {
            \input{snr_task2.pgf}};
        \node[label] at (9, 0) {\textbf{B}};
    \end{tikzpicture}

    \caption{%
    Reference experiments for the \acrfull{lcdtask} (\textbf{A}) and for the \acrfull{ssmftask} (\textbf{B}).
    The rows in \textbf{A} and \textbf{B} show the \gls{ber} (left column) and spiking activity (number of generated spikes, right column) over noise levels (upper row), number of hidden neurons in the network (middle row), and the number of neighboring samples $\ntap$ (last row).
    A lower noise power results in lower \gls{ber}.
    More hidden neurons show better demapping capabilities and a higher spiking activity.
    Because of less \gls{isi}, the \gls{lcdtask} requires fewer consecutive samples $\ntap$ than the \gls{ssmftask}.
    Experiments are based on \cite{arnold2023spiking,vonBank2024energy}.
    }
    \label{fig:task_results}
\end{figure*}

For both tasks, we train \gls{snn}-based receivers as references, optimized as discussed in Sec.~\ref{sec:symbol_level}.
The results are shown in Fig.~\ref{fig:task_results}.
The \glspl{snn} for the \gls{lcdtask} are based on \cite{arnold2023spiking}, while those for the \gls{ssmftask} are inspired by the Ternary Equalizer of~\cite{vonBank2024energy}.
Both tasks utilize the identical \gls{snn} architectures as well as input and output decoding schemes as given in the references.
For the \gls{lcdtask}, unless stated otherwise, we use $\ntap = 7$, $N^\text{h} = \num{40}$ hidden \gls{lif} neurons, and \gls{motm} decoding.
For the \gls{ssmftask}, the models use $\ntap = 21$, $N^\text{h} = \num{80}$ recurrently connected hidden \gls{lif} neurons, and \gls{eotm} decoding.
In the left columns in \cref{fig:task_results}A and B, we depict the \gls{ber}, while the right column shows the corresponding spike counts.
Input spike counts reflect the encoding of the input samples. The hidden spike counts correspond to the spikes within the network.
The first row illustrates the performance of the receiver at varying noise levels, where smaller noise levels (and thus higher \glspl{snr}) lead to more stable transmissions. 
The middle row investigates the influence of the hidden layer size and spike count at a fixed noise power of $\sigma_\mathrm{n}^2=-\SI{20}{\decibel}$.
Increasing the number of neurons reduces the \gls{ber}, but increases the network's spiking activity.
The final row varies $\ntap$.
For the \gls{lcdtask}, the \gls{ber} saturates at $\ntap = 7$, while for the \gls{ssmftask}, the \gls{ber} reaches a minimum at $\ntap=21$.
This indicates that the first task requires fewer consecutive samples to achieve a low \gls{ber} due to its lower \gls{isi} compared to the second task.
All models are tested on independent data until a minimum of \num{2000} bit errors are encountered.
The results are listed in \cref{tab:result_task1,tab:result_task2}.

\begin{table}[htb]
    \centering
    \caption{Results: \acrlong{lcdtask} at $\sigma_\mathrm{n}^2=-\SI{20}{\decibel}$}
    \label{tab:result_task1}
    \begin{tabular}{l cccc}
        \toprule
        \textbf{Reference} & \# \textbf{Param.} & \# \textbf{Spikes} & \textbf{BER} \\
        \midrule
        \cite{arnold2023spiking}, Sim. & 2960 & - & \num[round-precision=3,round-mode=figures,scientific-notation=true]{8.16935484e-04} \\
        \cite{arnold2023spiking}, \gls{bss2} & 2960 & - & \num[round-precision=3,round-mode=figures,scientific-notation=true]{0.0010035} \\[.5em]
        Examples: &  &  &  &   \\
        \quad $N^\text{h} = 5$ & 370 & 46.94 & \num[round-precision=3,round-mode=figures,scientific-notation=true]{0.0013753424657534246}  \\
        \quad $N^\text{h} = 40$ & 2960 & 77.82 & \num[round-precision=3,round-mode=figures,scientific-notation=true]{0.0006817567567567569}  \\
        \quad $N^\text{h} = 100$ & 7400 & 119.73 & \num[round-precision=3,round-mode=figures,scientific-notation=true]{0.0005626404494382024}  \\
        \bottomrule
    \end{tabular}
\end{table}

\begin{table}[htb]
    \centering
    \caption{Results: \acrlong{ssmftask} at $\sigma_\mathrm{n}^2=-\SI{20}{\decibel}$}
    \label{tab:result_task2}
    \begin{tabular}{l cccc}
        \toprule
        \textbf{Reference} & \# \textbf{Param.} & \# \textbf{Spikes} & \textbf{BER} \\
        \midrule
        \cite{vonBank2024energy}, Ternary Sim. & 32964 & 297.14 & \num[round-precision=3,round-mode=figures,scientific-notation=true]{0.000879} \\[.5em]
        Examples: &  &  &  &   \\
        \quad $N^\text{h} = 5$ & 889 & 84.96 & \num[round-precision=3,round-mode=figures,scientific-notation=true]{0.02029}  \\
        \quad $N^\text{h} = 40$ & 8484 & 121.47 & \num[round-precision=3,round-mode=figures,scientific-notation=true]{0.00162823}  \\
        \quad $N^\text{h} = 80$ & 20164 & 212.3 & \num[round-precision=3,round-mode=figures,scientific-notation=true]{0.00105526}  \\
        \quad $N^\text{h} = 100$ & 27204 & 269.92 & \num[round-precision=3,round-mode=figures,scientific-notation=true]{0.00104896}  \\
        \bottomrule
    \end{tabular}
\end{table}

%% file: dataset.tex
\section{Code and Dataset}
\begin{figure*}[tbh]
    \centering
    \tikzset{
        panel/.style={
            inner sep=0pt,
            outer sep=0pt,
            execute at begin node={\tikzset{anchor=center, inner sep=.33333em}}},
        label/.style={
            anchor=north west,
            inner sep=0,
            outer sep=0}}

    \begin{tikzpicture}
        \node[panel, anchor=north west] (b) at (0,  0) {
            \input{shapes.tex}};
    \end{tikzpicture}

    \caption{Visualization of the tensors returned by the provided datasets.
    A sequence of bit pairs $\bm{B}$ is mapped to symbols in $\mathcal{X}$ via indices \texttt{q}.
    The symbols are transmitted, and at the receiver $\bm{y}$ is observed.
    Those symbols are reshaped into overlapping chunks \texttt{y\_chunk} containing neighboring samples to capture the information spread because of \gls{isi}, needed to demap \texttt{y[k]}.
    The receiver infers bit decisions $\hat{b}_1[k]\hat{b}_2[k]$. 
    }
    \label{fig:imdd_link}
\end{figure*}
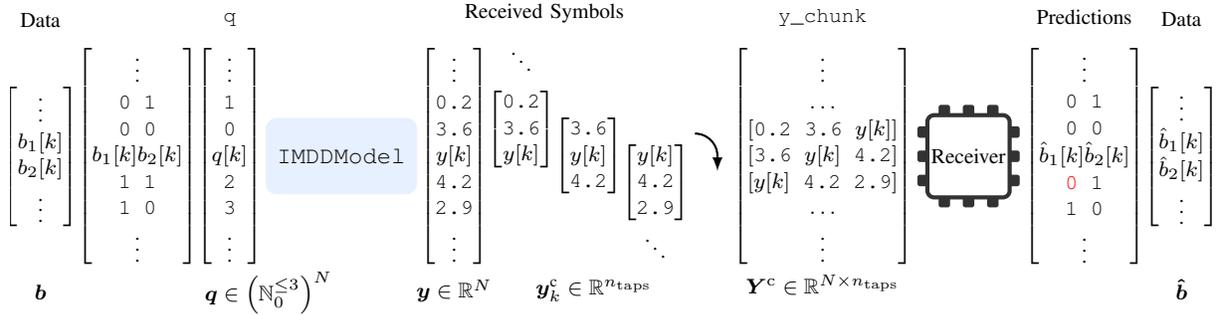

The implementation of the \gls{imdd} link is publicly available at \href{https://github.com/imdd-task/imdd-task}{github.com/imdd-task/imdd-task}.
We provide a PyTorch-based \texttt{Dataset}, enabling seamless integration with existing \gls{snn} training frameworks such as Norse~\cite{pehle2021norse}.
The datasets corresponding to the defined tasks are imported as \texttt{LCDDataset} and \texttt{SSMFDataset} and contain link parameterizations as defined in Sec.~\ref{sec:task}.
An example usage is shown in Sec.~\ref{lst:dataset}.
Notably, after completing one epoch (i.e., accessing all $\texttt{N}$ samples in the dataset), the dataset generates new data instead of reusing the same transmit samples, which can be disabled by setting \texttt{continuous\_sampling=False}.
The number of taps $\ntap$ and the noise power $\sigma_\text{n}^2$ are set by via \texttt{set\_n\_taps} and \texttt{set\_noise\_power\_db}.
\begin{lstlisting}[
    caption={Loading the IM/DD dataset.},
    captionpos=b,
    label={lst:dataset},
    basicstyle=\tt\footnotesize,
    escapeinside=\'\',
    numbers=left,
    xleftmargin=2em,
    numberstyle=\footnotesize\color{black!30},
    stepnumber=1]
'\tt\color{color0}from' torch.utils.data '\tt\color{color0}import' DataLoader
'\tt\color{color0}from' IMDD '\tt\color{color0}import' LCDDataset
'\tt\color{gray}\emph{\# Dataset}'
dataset = '\tt\color{color1}LCDDataset'(bit_wise=False)
dataset.set_n_tap(n_tap)
'\tt\color{gray}\emph{\# Data loader}'
dataloader = '\tt\color{color1}DataLoader'(
    dataset, batch_size, shuffle=True)
for (y_chunk, q) in dataloader:
    ... '\tt\color{gray}\emph{\# train}'
\end{lstlisting}

Internally, the datasets utilize an instance of \texttt{IMDDModel}, which simulates the link as detailed in Sec.~\ref{sec:imddmodel}.
The model generates indices $q[k]$ of transmit symbols $\boldsymbol{x}$ in $\mathcal{X}$ and outputs the received chunks $\boldsymbol{y}^\text{c}[k]$, from which the pairs of bits $\hat{b}_1[k]\hat{b}_2[k]$ are estimated.
The indices \texttt{q} of sent symbols are of shape \texttt{(N,)}.
Each transmit symbol is labeled with its assigned bits, generated using the provided \texttt{get\_graylabels} function, and accessed at the corresponding index \texttt{q}.
By default, the dataset does not directly return bit-level labels to maintain flexibility.
This design supports symbol-level receivers, allowing models to output class index $\hat{q}$ corresponding to the predicted symbol $\hat{x} \in \mathcal{X}$, which can be used together with $q$ in a cross-entropy loss function.
For tasks requiring bit-level outputs, setting \texttt{bit\_level=True} in the dataset configuration changes the format of $\bm{q}$ to \texttt{(N, 2)}, where each entry contains binary values corresponding to the respective bits.
The chunk tensor has shape \texttt{(N, n\_taps)}, representing the chunked received data.
Hence, at each index $k$, the tensor holds all symbols required to process $y[k]$ at position \texttt{(k, n\_taps//2)}.
This allows shuffling and batching along the first dimension using a PyTorch \texttt{Dataloader}, resulting in \texttt{y\_chunk} of shape \texttt{(batch\_size, n\_taps)} and \texttt{q} of shape \texttt{(batch\_size,)}. 
To compute the \gls{ber} based on either predicted symbols or bits, we provide a helper function \texttt{bit\_error\_rate}.
Fig.~\ref{fig:imdd_link} visualizes the tensors.

An arbitrarily parameterized \gls{imdd} link can be created by passing an instance of \texttt{IMDDParams} to an \texttt{IMDDModel}.
A dataset is created by instantiating \texttt{IMDDDataset} with the given parameters, see \cref{lst:imdd_link_general}.
This allows users to adapt the provided link implementations to suit their specific requirements.
\begin{lstlisting}[
    caption={Creating an arbitrarily parameterized IM/DD link and dataset.},
    captionpos=b,
    label={lst:imdd_link_general},
    basicstyle=\tt\footnotesize,
    escapeinside=\'\',
    numbers=left,
    xleftmargin=2em,
    numberstyle=\footnotesize\color{black!30},
    stepnumber=1]
'\tt\color{color0}from' IMDD '\tt\color{color0}import' IMDDDataset, IMDDParams
'\tt\color{gray}\emph{\# Parameterization}'
params = '\tt\color{color1}IMDDParams'(...)
'\tt\color{gray}\emph{\# Link model}'
link = '\tt\color{color1}IMDDModel'(params)
'\tt\color{gray}\emph{\# Dataset}'
dataset = '\tt\color{color1}IMDDDataset'(params, bit_level=False)
\end{lstlisting}
The code to generate the data in \cref{fig:imdd_link_data} and \cref{fig:task_results}, and documented examples are provided in our public repository.

%% file: shapes.tex
\begin{tikzpicture}
    \tikzset{
    font={\fontsize{8pt}{10}\selectfont},
    arr/.style = {draw=black, rounded corners=3mm, thick, -latex},
    device/.style={
        draw,
        rounded corners=1pt,
        font=\scriptsize},
    outer/.style={
        draw,
        dotted,
        rounded corners=2pt,
        anchor=south west},
    tx/.style={
        minimum width = 5.9cm,
        minimum height = 1.5cm},
    rx/.style={
        minimum width = 7.58cm,
        minimum height = 1.5cm}}

    \node (bvec) at (0, 0) {$
    \begin{bmatrix}
        \vdots \\
        b_1[k] \\
        b_2[k] \\
        \vdots
    \end{bmatrix}
    $};
    \node[above=0.5cm of bvec] (data) {Data};
    \node[below=0.5cm of bvec] (b) {$\boldsymbol{b}$};

    \node[right=-7px of bvec] (Bvec) {$
    \begin{bmatrix}
        \vdots \\
        \texttt{0 1} \\
        \texttt{0 0} \\
        b_1[k]b_2[k] \\
        \texttt{1 1} \\
        \texttt{1 0} \\
        \vdots
    \end{bmatrix}
    $};

    \node[right=-7px of Bvec] (qvec) {$
    \begin{bmatrix}
        \vdots \\
        \texttt{1} \\
        \texttt{0} \\
        q[k] \\
        \texttt{2} \\
        \texttt{3} \\
        \vdots
    \end{bmatrix}
    $};
    \node[above=0.0cm of qvec] (q) {\texttt{q}};
    \node[below right=-0.2cm and -1cm of qvec] (Q) {$\boldsymbol{q} \in \left(\mathbb{N}_0^{\leq 3}\right)^{N}$};

    \draw[rounded corners, fill=color1!10, draw=color1!10] (3.0, -0.5) rectangle node{\small{\texttt{IMDDModel}}} (5, 0.5);

    \node (Yvec) at (5.5, 0) {$
    \begin{bmatrix}
        \vdots \\
        \texttt{0.2} \\
        \texttt{3.6} \\
        y[k]  \\
        \texttt{4.2} \\
        \texttt{2.9} \\
        \vdots
    \end{bmatrix}
    $};
    \node[below=-0.1cm of Yvec] (Y) {$\boldsymbol{y} \in \mathbb{R}^{N}$};

    \node () at (6.7, 1.9) {Received Symbols}; 
    \node[above right=-54px and -7px of Yvec] (Yvec1) {$
    \begin{bmatrix}
        \texttt{0.2} \\
        \texttt{3.6} \\
        y[k]
    \end{bmatrix}
    $};
    \node[above of=Yvec1] {$\ddots$};
    \node[below right=-26.5px and -7px of Yvec1] (Yvec2) {$
    \begin{bmatrix}
        \texttt{3.6} \\
        y[k] \\
        \texttt{4.2}
    \end{bmatrix}
    $};
    \node[right=10px of Y] {$\boldsymbol{y}_k^\text{c} \in \mathbb{R}^{n_\mathrm{taps}}$}; 
    \node[below right=-26.5px and -7px of Yvec2] (Yvec3) {$
    \begin{bmatrix}
        y[k] \\
        \texttt{4.2} \\
        \texttt{2.9} \\
    \end{bmatrix}
    $};
    \node[below=-7px of Yvec3] {$\ddots$};

    \draw[arr]  (8.7, 0.3)  -|  (9, -0.2); 

    \node (Ych) at (10.4, 0) {$
    \begin{bmatrix}
        \vdots \\
        \dots \\
        \left[\texttt{0.2  } \texttt{3.6  } y[k]\right] \\  
        \left[\texttt{3.6  } y[k] \texttt{  4.2}\right] \\  
        \left[y[k] \texttt{  4.2} \texttt{  2.9}\right] \\  
        \dots \\
        \vdots
    \end{bmatrix}
    $};
    \node[above=0.0cm of Ych] {\texttt{y\_chunk}};
    \node[below=-4.5px of Ych] {$\bm{Y}^\mathrm{c} \in \mathbb{R}^{N\times n_\mathrm{taps}}$};

    \node[right=-5px of Ych] (chip) {\resizebox{.075\textwidth}{!}{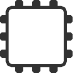}};
    \node (rec) at (chip.center) {\footnotesize Receiver};
    \node[right=-5px of chip] (Bhatvec) {$
    \begin{bmatrix}
        \vdots \\
        \texttt{0 1} \\
        \texttt{0 0} \\
        \hat{b}_1[k]\hat{b}_2[k] \\
        \texttt{\textcolor{red}{0} 1} \\
        \texttt{1 0} \\
        \vdots
    \end{bmatrix}
    $};
    \node[above=0.0cm of Bhatvec] (Bhat) {Predictions};

    \node[right=-7px of Bhatvec] (bhatvec)  {$
    \begin{bmatrix}
        \vdots \\
        \hat{b}_1[k] \\
        \hat{b}_2[k] \\
        \vdots
    \end{bmatrix}
    $};
    \node[above=0.5cm of bhatvec] {Data};
    \node[below=0.4cm of bhatvec] {$\boldsymbol{\hat{b}}$};
\end{tikzpicture}

%% file: acknowledgments.tex
\section*{Acknowledgements}
\addcontentsline{toc}{section}{Acknowledgment}

This work has received funding from
the EC Horizon 2020 Framework Programme
under grant agreements
945539 (HBP SGA3), 
the EC Horizon Europe Framework Programme
under grant agreement
101147319 (EBRAINS 2.0),
the Deutsche Forschungsgemeinschaft (DFG, German Research Foundation) under Germany's Excellence Strategy EX 2181/1-390900948 (the Heidelberg STRUCTURES Excellence Cluster).
This work has received funding from the European Research Council (ERC) under the European Union's Horizon 2020 research and innovation program (grant agreement No. 101001899).